\documentclass[sn-mathphys]{sn-jnl}%

\usepackage{siunitx}
\jyear{2022}%
\theoremstyle{thmstyleone}%
\theoremstyle{thmstyletwo}%
\theoremstyle{thmstylethree}%
\begin{document}
\title[ ]{The secret role of undesired physical effects in accurate shape sensing with FBGs}
\author*[1]{\fnm{Samaneh} \sur{Manavi Roodsari}}\email{samaneh.manavi@unibas.ch}
\author[1]{\fnm{Sara} \sur{Freund}}
\author[2]{\fnm{Martin} \sur{Angelmahr}}
\author[1]{\fnm{Georg} \sur{Rauter}}
\author[2]{\fnm{Wolfgang} \sur{Schade}}
\author[1]{\fnm{Philippe} \sur{C. Cattin}}
\affil[1]{\orgdiv{Department of Biomedical Engineering}, \orgname{University of Basel}, \orgaddress{\street{Gewerbestrasse 14}, \city{Allschwil}, \postcode{4123}, \country{Switzerland}}}
\affil[2]{\orgdiv{Department of Fiber Optical Sensor Systems}, \orgname{Fraunhofer Institute for Telecommunications, Heinrich Hertz Institute, HHI}, \orgaddress{\street{Am Stollen 19H}, \city{Goslar}, \postcode{38640}, \country{Germany}}}
\abstract{
Fiber optic shape sensing is an innovative technology that has enabled remarkable advances in various navigation and tracking applications. Although the state-of-the-art fiber optic shape sensing mechanisms can provide sub-millimeter spatial resolution for off-axis strain measurement and reconstruct the sensor's shape with high tip accuracy, their overall cost is very high. The major challenge in more cost-effective fiber sensor alternatives for providing accurate shape measurement is the limited sensing resolution in detecting the shape deformations. 
Here, we present a novel data-driven technique to overcome this limitation by removing strain measurement, curvature calculation, and shape reconstruction steps. We design a deep-learning model based on convolutional neural networks that is trained to directly predict the sensor's shape based on its spectrum. Our fiber sensor is based on easy-to-fabricate eccentric fiber Bragg gratings (FBG) and is interrogated with a simple and cost-effective readout unit in the spectral domain. We demonstrate that our deep-learning model benefits from undesired bending-induced effects (\textit{e.g.,} cladding mode coupling and polarization), which contain high-resolution shape deformation information. These findings are the first steps toward a low-cost yet accurate fiber shape sensing solution for detecting complex multi-bend deformations.
}


\maketitle

Fiber optic shape sensing has proven to have great potential, specifically in medical applications such as catheter navigation, surgical needle tracking, and continuum robot navigation. Compared to other common navigation technologies~\cite{OpticalTracking,OpticalTracking_2,wagner20164d}, fiber shape sensing has many advantages, such as immunity to electromagnetic fields, bio-compatibility, and high flexibility. Fiber shape sensors are small in diameter, easily integrable into flexible instruments, and require no line-of-sight. Distributed sensors based on multicore fibers can also provide high-resolution shape measurements~\cite{soller2005polarization,TSSC,meng2021shape}.

Fiber shape sensors measure off-axis strain, which is then used to calculate the directional curvature and reconstruct the sensor's shape~\cite{marowsky2014planar}. Various fiber sensor configurations have been investigated for off-axis strain measurement, including multicore fibers with~\cite{khan2019multi,moore2012shape,bronnikov2019durable} or without~\cite{nishio2008shape,zhao2016distributed,issatayeva2021design} FBGs in their cores, fibers with cladding waveguide FBGs~\cite{waltermann2015cladding}, and fiber bundles made from multiple single-mode fibers that contain FBG arrays~\cite{roodsari2022fabrication,manavi2018temperature,moon2015fbg,ryu2014fbg,roesthuis2013using}.
For an accurate shape reconstruction, high spatial resolution for off-axis strain measurement is essential. In some fiber shape sensor configurations (\textit{e.g.,} distributed multicore fiber sensor), sub-millimeter spatial resolution can be achieved~\cite{soller2005polarization}. However, in these sensors, complex and expensive readout units are used to analyze the output signal in time- or frequency domain for strain measurement~\cite{eickhoff1981optical,masoudi2016contributed,bao2012recent,yuksel2018rayleigh}. Although fiber sensors interrogated with the spectral-domain readout systems are cheaper, their spatial resolution is limited by their lower sensing plane density~\cite{khan2019multi,beisenova2019distributed}, making them inapplicable for tracking complex shape deformations. Therefore, a cost-effective, high-resolution, and accurate fiber shape sensing technique is desirable.

\begin{figure}[!t]
    \centering
    \includegraphics[width=1\textwidth]{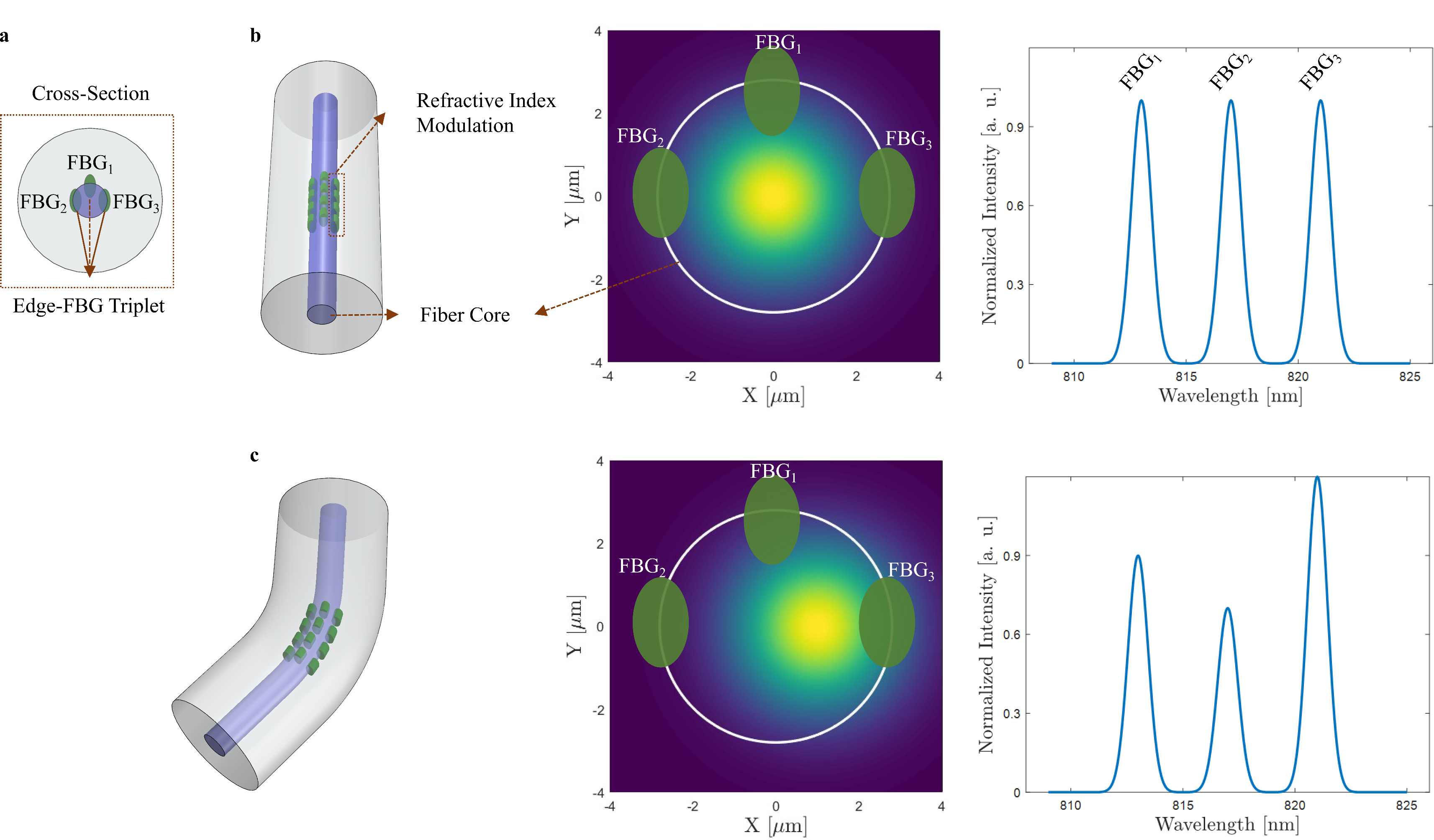}
    \caption{\textbf{FBG configuration and working principle of the eFBG sensor.} \textbf{a} Sketch of the cross-section view of the eFGB sensor. Each sensing plane of the eFBG sensor consists of three FBGs inscribed off-axis with  $\sim$\,$90^{\circ}$ angular separation (also known as edge-FBG triplet). \textbf{b} Mode-field distribution of a straight single-mode fiber and the expected signal from eFBGs of a same sensing plane. \textbf{c} When the fiber is curved, mode-field distribution moves in the opposite direction of the bending, which affects the relative intensity between the eFBGs.}
    \label{fig:mode_field}
\end{figure}

Among cost-effective fiber shape sensors interrogated in the spectral domain, eccentric FBG (eFBG) sensors show great capacity for tracking applications thanks to their unique sensing mechanism~\cite{waltermann2018multiple,bao2018all,rong2017highly}. Each sensing plane in eFBG shape sensors consists of three highly localized FBGs, written off-axis in the fiber's core (also known as edge-FBG triplet) as shown in Fig.~\ref{fig:mode_field}a~\cite{waltermann2018multiple}. 
Shape deformations are commonly calculated from the displacement of the fundamental mode-field inside the optical fiber, which is estimated from spectral intensity modifications (See Figs.~\ref{fig:mode_field}b and c)~\cite{waltermann2018multiple,bao2018all}. However, many other effects including bending-sensitive mode coupling~\cite{thomas2011cladding,thomas2012cladding,erdogan1997cladding,feng2016off}, polarization-dependent losses~\cite{galtarossa2005polarization,smith1980birefringence,ulrich1980bending,kersey1997fiber,block2006bending}, and wavelength-dependent bending losses~\cite{marcuse1976field,faustini1997bend,valiente1989new,harris1986bend,murakami1978bending,morgan1990temperature,renner1992bending} also modify the eFBGs spectral profile. These effects cannot be accurately modeled, and their impact on the sensor's spectra is indistinguishable from the mode-field displacements. Further details on the eFBG configuration, sensing mechanism, and bending-induced effects are provided in Methods.

We present in this paper a data-driven modeling technique based on deep learning (DL) that can indeed find a meaningful pattern in the eFBG signal that is affected by uncontrolled bending-induced effects. These additional sources of information considerably improve shape prediction accuracy. Our novel technique provides high spatial resolution shape estimation, directly from the eFBG sensor's signal without requiring any strain measurement, curvature calculation, and shape reconstruction steps.
%
%
%
\subsection*{Concept}\label{Concept}
 \begin{figure}[!b]
    \centering
    \includegraphics[width=0.85\textwidth]{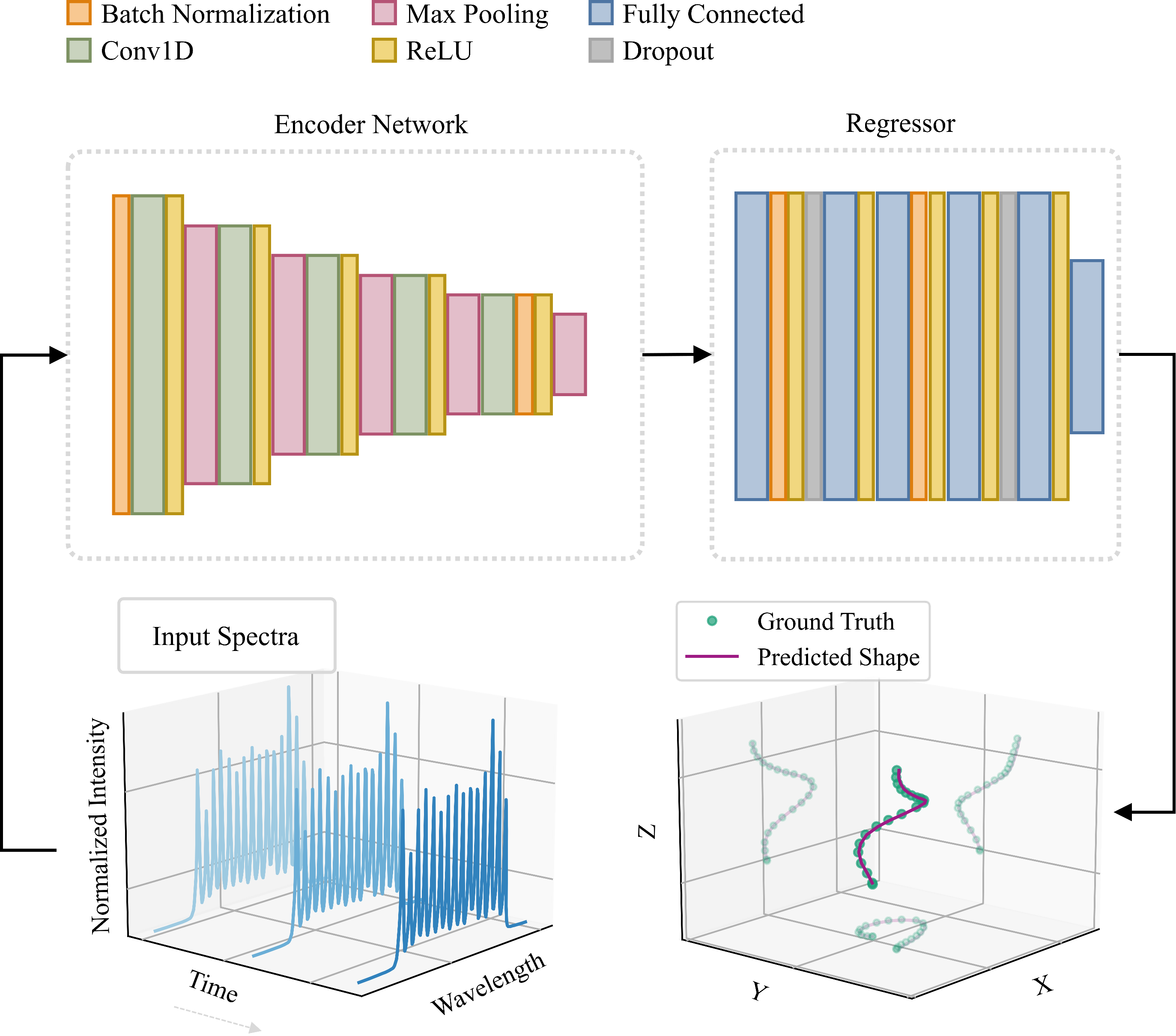}
    \caption{\textbf{Architecture of the best-performing configuration after hyperparameter tuning}. The architecture includes five 1D convolutional layers (Conv1D), six fully connected layers, five max pooling layers, four batch normalization steps, and two dropout steps. The designed network receives three consecutive spectral scans as the input and predicts the relative coordinates of 20 discrete points over the sensor's curve. More detail on the channel, kernel, and pooling sizes is available under Methods.}
    \label{fig:architecture}
\end{figure}

In this section, we explain the designing and training process of a deep neural network for our eFBG shape sensor. 
The \SI{30}{\cm} long eFBG fiber sensor used in this work features five sensing planes separated by \SI{5}{\cm} from each other. At each sensing plane, three off-axis FBGs are inscribed at a radial distance of $\sim$\,\SI{2}{\micro\m} to the top, left, and right side of the fiber's core. 

The dataset used for developing the deep-learning-based model is collected using a similar setup reported in our previous work~\cite{manavi2021using} (see Methods for more details). We use three normalized, consecutively measured spectral scans as input data to the proposed DL model. Each scan is recorded from \SIrange{800}{890}{\nm} comprising 190 wavelength components. The target data are the relative coordinates of 20 discrete points (reflective markers of the tracking system) measured over the length of the shape sensor (see Ref.~\cite{manavi2022usingarxive} for more detail on data preprocessing). For this dataset, around 58000 samples are collected during 30\,mintues of random movement of the fiber sensor. To evaluate the predictive performance of the trained model in an unbiased way, samples are first shuffled and then split into Train-Validation-Test subsets: 80$\%$ for training, 10$\%$ for validating, and 10$\%$ for testing. We refer to this testing data set as $Test_1$ for the remainder of this paper. The second set of data ($Test_2$) with a size of $\sim$\,5800 samples is recorded separately to evaluate the performance of the trained model for unseen shapes from a continuous movement. We also collected 320 samples, as $Test_3$, when only certain sensor regions are bent (see Methods for more detail).

A DL model needs a specially designed network architecture to extract essential features from the sensor's spectra and to predict its corresponding shape. To do so, we ran an optimization algorithm similar to the Hyperband optimizer~\cite{li2017hyperband}. This optimization algorithm looks for the best set of essential parameters, such as number of layers, whose values can not be estimated from the data during training (also known as hyperparameters). Figure~\ref{fig:architecture} shows the architecture of the best-performing configuration after hyperparameter tuning. 

To find out which part of the spectra is relevant for feature extraction, we calculate the forward finite difference of the network's output with respect to the input spectral components. This difference provides an influence evaluation for each wavelength component of the input spectra to decode the model's predictions (see Methods).

As an evaluation baseline, we compared the shape prediction accuracy of the proposed DL approach with the mode-field dislocation method (MFD) on the same test sets. Following the process explained in our previous work~\cite{waltermann2018multiple}, we calibrate our shape sensor to determine the exact angular and radial position of each eFBG. Then, we estimate the mode-field centroid at each sensing plane and calculate the curvature and the bending direction~\cite{waltermann2018multiple}. Finally, we reconstruct the 3D shape of the eFBG sensor using the interpolated values of the calculated directional curvatures at small arc elements. It should be noted that the density of the sensing planes in our eFBG shape sensor is not sufficient for the MFD method to estimate complex deformations. Nevertheless, we performed the test to show the superiority of the proposed data-driven technique (DL).

\subsection*{Results and Discussion}\label{}
\textbf{Shape prediction evaluation}. 
We evaluated the performance of the DL approach using the three testing datasets and compared the results with the MFD method. Table~\ref{tab:shapeevaluation} shows the shape error metrics including the tip error, that is, the Euclidean distance between the true and the predicted coordinate of the sensor's tip, and the root-mean-square of the Euclidean distance (RMSE) between the true and the predicted coordinates of the discrete points along the sensor's length.
The MFD approach, when using $Test_1$ dataset, shows median and interquartile (IQR) tip error values of \SI{111.3}{\mm} and \SI{121.5}{\mm}, respectively. These error values reduce to \SI{98.5}{\mm} and \SI{46}{\mm} when using $Test_2$ dataset. The reason for such performance difference is that $Test_1$ dataset contains more diverse shapes, as the samples are randomly selected from a larger dataset compared to $Test_2$, which is a continuous sensor movement in a short period. As expected, the error values are considerably high in all testing datasets, since there is too little information available for the MFD approach to estimate the complex shape deformations in these datasets.
\begin{table}[!t]
    \centering
    \fontsize{8}{10}\selectfont
    \caption{\textbf{Shape evaluation errors in MFD and DL methods using test sets $\boldsymbol{Test_1}$, $\boldsymbol{Test_2}$, and $\boldsymbol{Test_3}$}. The lowest achieved error values are indicated in bold.}
    \label{tab:shapeevaluation}
    \begin{tabular}{cccccc}
    \hline

\multicolumn{2}{c}{\textbf{ }} & \multicolumn{2}{c}{\textbf{tip error\,{[}mm{]}}} & \multicolumn{2}{c}{\textbf{RMSE\,{[}mm{]}}} \\ 
    \text{dataset}                   &\text{method} &\text{median} &\text{IQR} &\text{median} &\text{IQR} \\ \hline
    \multirow{2}{*}{\textbf{$Test_1$}}  &\text{MFD}  & 111.3 & 121.5 & 59.4  & 71.7\\
                                        &\text{DL}  &\textbf{2.1}   &\textbf{2.6}   &\textbf{1.5}   &\textbf{1.6}\\
    \multirow{2}{*}{\textbf{$Test_2$}}  &\text{MFD}  &98.5   &46.0   &53.8   &29.1\\
                                        &\text{DL}  & 17.1  & 12.6  & 9.8   & 7.0\\
    \multirow{2}{*}{\textbf{$Test_3$}}  &\text{MFD}  & 39.5  & 34.7  & 17.1  & 18.3\\
                                        &\text{DL}  &6.0  &9.0  &5.1  &6.6\\ \hline
    \end{tabular}
\end{table}

The DL method, on the other hand, significantly improves the shape prediction accuracy of $Test_1$ samples, with a median and IQR tip error values of \SI{2.1}{\mm} and \SI{2.6}{\mm}. These values increase to \SI{17.1}{\mm} and \SI{12.6}{\mm} on less diverse $Test_2$ samples. This is due to the fact that a DL model can only learn to extract the most general/relevant features from the input signal, if the training dataset is representative of the expected signals from the sensor. However, in $Test_2$ dataset, less than 2\% of the samples have at least 100 similar examples in the training data (a maximum RMSE of \SI{5}{\mm} is chosen as the similarity measure after evaluating several thresholds). This shows that 30\,minutes of manual shape manipulation is insufficient to cover the sensor's working space and create a representative training dataset for the model to generalize properly. On the other hand, in $Test_1$ dataset, almost 20\% of the samples have at least 100 similar examples in the training dataset, which means the DL method is being tested on samples that the model has already learned how to handle. Therefore, $Test_1$ can mimic the situation where the training dataset represents the expected shapes of the sensor.

Shape evaluation results of $Test_1$ dataset define the performance of our model's lower limit. Such performance difference also suggests that the DL model is better to be trained as application-specific, since it can better focus on relevant features when learned from the expected shape distribution of the sensor. On the other hand, when training data covers most of the expected behaviors from the sensor, the DL model might only ``memorize'' the corresponding shape for each signal without searching for relevant features in the sensor's spectrum. To investigate this, we compare the performance of our DL method with a dictionary-based algorithm. In this approach, all the training and validation samples create a pre-defined dictionary. The shape prediction is made by looking for the closest spectrum to the test sample and presenting its corresponding shape. The median tip errors on datasets $Test_1$ and $Test_2$ are \SI{5.9}{\mm} and \SI{50.0}{\mm} with IQR values of \SI{3.9}{\mm} and \SI{43.3}{\mm}, respectively, which are higher than error values when using our DL technique. This shows that our DL model generalizes and is indeed beneficial for predicting more accurate shapes.

Two essential factors have to be considered when working with dictionaries: the size of the dictionary and the execution time required to find the best matching example. To get an accurate shape estimation for a given sample, the number of stored samples in the dictionary should be large enough to cover all possible examples, which leads to a long execution time. Therefore, this approach has a trade-off between accuracy and execution time. However, extensive training data do not negatively affect these two factors in DL method, as the resulting model size is independent of the training data size.

\begin{figure}[!b]
    \centering
    \includegraphics[width=1\textwidth]{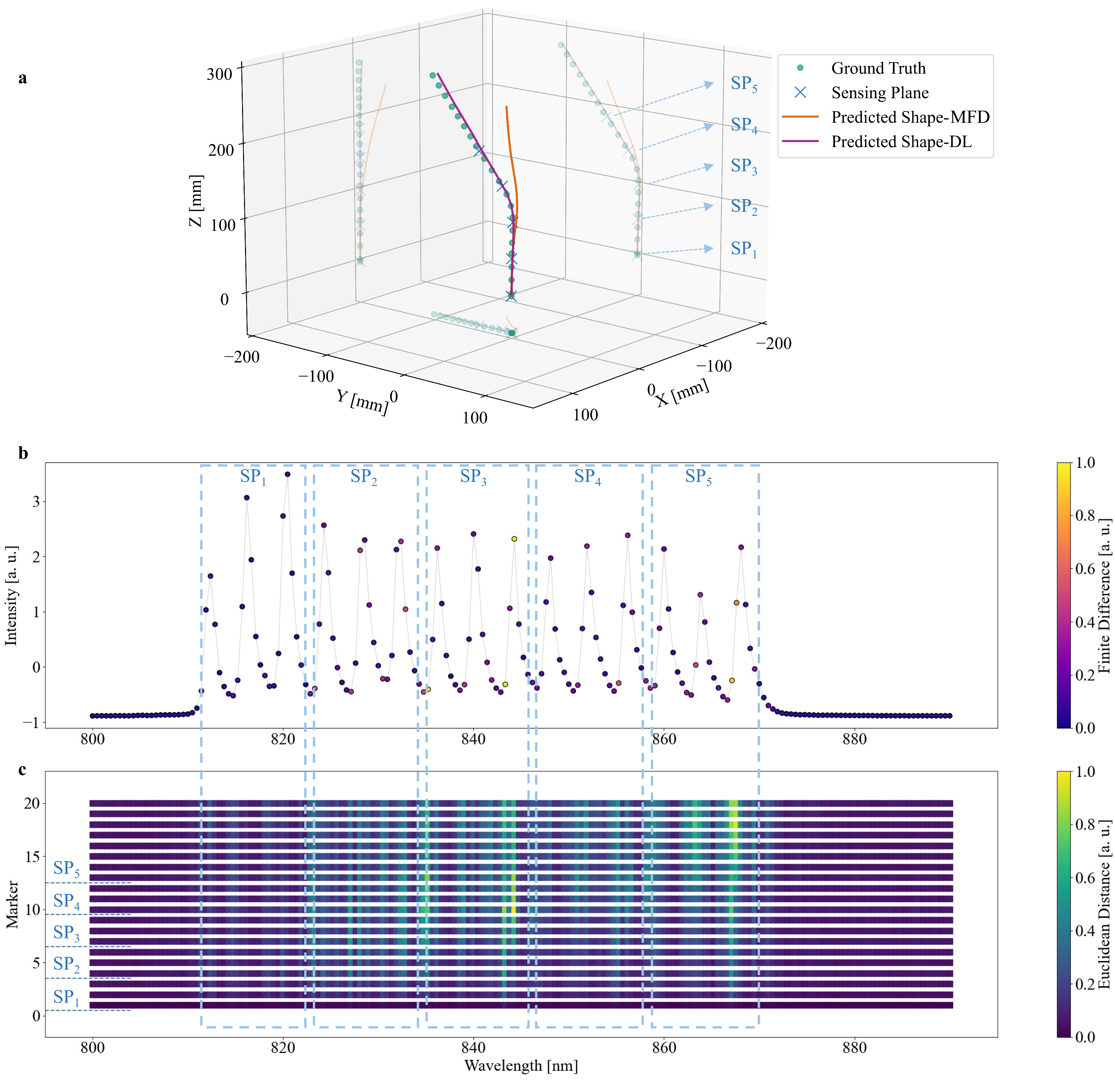}
    \caption{\textbf{Decoding the DL model decision for deformations between sensing planes}. \textbf{a} An example from $Test_3$ samples in which the sensor is bent in the region between the sensing planes 3 and 4. The true shape (ground truth) is shown with green circles. The five sensing planes of the sensor are shown with $\times$ signs. The predicted shapes using the mode-field dislocation method (MFD) and the deep learning method (DL) are shown with orange and purple solid lines, respectively. \textbf{b} The finite difference of the loss value with respect to the input spectral elements. Wavelength components shown with colors closer to yellow contribute more to the model's decision on this particular example. \textbf{c} Highlighting the importance of input spectral elements in relative coordinate prediction of all 20 markers based on the magnitude of the Euclidean distance between the predicted relative coordinates of each marker, before and after spectral modification. The position of the sensing planes with respect to the markers are indicated with dashed blue lines. SP$_i$: \textit{i}$_{\text{th}}$ Sensing Plane.}
    \label{fig:750}
\end{figure}

Our observations show that the designed DL model can recognize deformations even between the sensing planes. To further investigate this interesting finding, we evaluated the shape predictions using $Test_3$ dataset, in which the deformations are only applied between the sensing planes. $Test_3$ dataset contains four deformation examples, each repeated twice and measured 40 times.  As expected, the classical MFD method is not able to accurately predict the sensor's shape for such deformations, as the deformed area is not at any of the sensing planes. However, when using the DL method, we achieve a median tip error of \SI{6}{\mm} which is $\sim$ six times smaller than the median tip error using MFD on this dataset. The precision of the predicted tip position in $Test_3$ dataset is \SI{1.9}{\mm} on average. 

An example from $Test_3$ samples where the sensor is bent in the region between the sensing planes 3 and 4 is depicted in Fig.~\ref{fig:750}a. It should be noted that the intensity ratio of the eFBG Bragg peaks in each sensing plane can also be influenced by previously mentioned effects other than fundamental mode-field dislocations. The MFD approach, however, does not consider such effects and is thus incapable of correctly interpreting the signal variations. On the other hand, the DL model manages to accurately predict the sensor's shape as it looks at the full spectral profile, including the minute changes at the wavelengths outside the Bragg resonances. Figure~\ref{fig:750}b shows the finite difference analysis of the loss value with respect to the 190 wavelength components of the input spectra. The higher the difference, the more important the corresponding wavelength component is for shape prediction in this example. Figure~\ref{fig:750}c gives a deeper insight into this investigation. For all 190 wavelength components, the Euclidean distance between the predicted relative coordinates of each marker before and after the spectral modification is depicted using a color map. The contribution of each wavelength component to the relative coordinate prediction of all 20 markers is realized from the presented color map in Fig.~\ref{fig:750}c.

\begin{figure}[!b]
    \centering
    \includegraphics[width=1\textwidth]{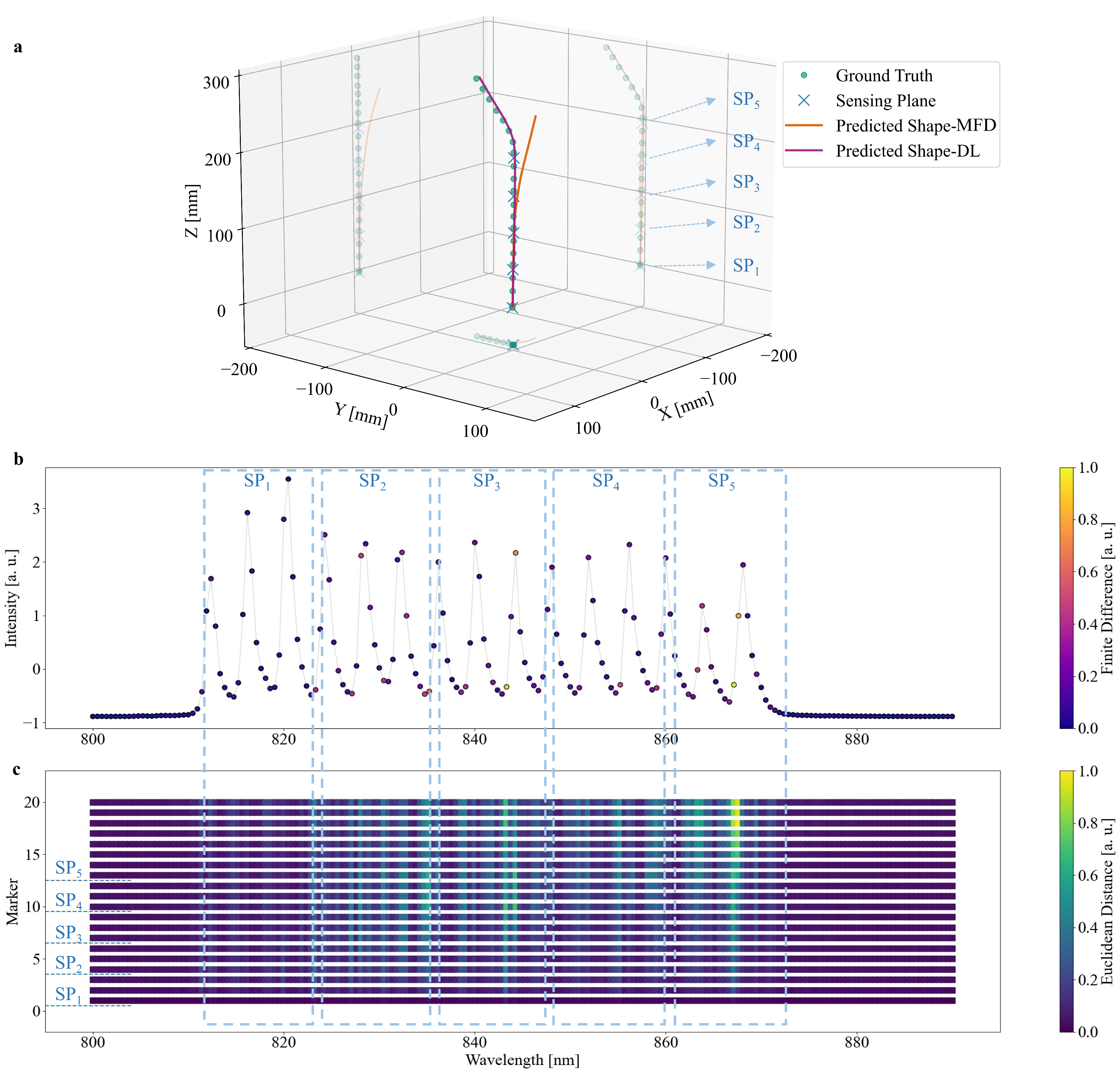}
    \caption{\textbf{Decoding the DL model decision for deformations after the last sensing plane}. An example from $Test_3$ samples in which a \SI{3}{\cm} long segment, \SI{1}{\cm} after the last sensing plane, is deformed. Refer to the caption of Fig.~\ref{fig:750} for more details.}
    \label{fig:1320}
\end{figure}

Another important finding is that the DL model can also detect deformations after the last sensing plane. Figure~\ref{fig:1320} shows an example in which, a \SI{3}{\cm} long segment, \SI{1}{\cm} after the last sensing plane, is deformed. Similar to the example in Fig.~\ref{fig:750}, the MFD method is not able to predict the sensor's shape in such deformations. The DL model, in contrast, learned to employ relevant features in the side slopes of the eFBG spectra to predict the correct shape (see Figs.~\ref{fig:1320}b and c). A possible explanation for such intriguing performance is that in the area after the last sensing plane, wavelength-dependent interference occurs between the back-reflected light from the air-glass interface at the fiber's end tip (Fresnel reflection) and the downstream incident light. Deformations in this region affect interferences in two ways: first, the spectral profile of the downstream light changes due to the bending. Second, the coupling conditions between the back-reflected and the downstream lights change. Consequently, the measured spectra from the fiber sensor show small variations, as the deformations affect the interference pattern.\\
\textbf{Optimum number of sensing planes}. 
A key factor in eFBG sensors, when using the MFD method, is the number of sensing planes for detecting shape deformations. Similar to any other quasi-distributed shape sensor, the distance between the sensing planes determines the sensor's spatial resolution in shape measurements. Depending on the complexity of the shape deformations, a limited number of sensing planes in the sensor (low spatial resolution) can lead to large tip errors in methods that include shape reconstruction (\textit{e.g.,} the MFD method). In this section, we present a theoretical analysis for realizing the minimum number of sensing planes required in eFBG sensors when using the MFD method to reach the same accuracy for the shape prediction as we get using five sensing planes in our DL method here.

We simulated the shape reconstruction error for different spatial resolutions. To do so, we first interpolate the discrete curve points over the sensor's true shape measured by the motion capture system, using a Spline with a resolution of \SI{0.1}{\mm} (this value was selected empirically). We then calculated the curvature and the torsion--the curve's deviation from the osculating plane--at the query points. Finally, we use the calculated curvatures and bending directions at the sensing planes to reconstruct the spatial curve and compare it with the true shape. For a \SI{25}{\cm} long sensor with \SI{50}{\mm} spatial resolution (five sensing planes), the median tip error of the reconstructed shapes, tested on $Test_1$ and $Test_2$ datasets, is $\sim$\,\SI{50}{\mm}, which is almost 16 times higher compared to what the DL approach achieved (see Table~\ref{tab:shapeevaluation}). In order to get a median tip error of \SI{3}{\mm}, the spatial resolution of the sensor should also be in a similar range, meaning that the MFD method would need around 84 sensing planes consisting of 252 eFBGs.

\subsection*{Conclusion}\label{}

In this paper, we developed a novel fiber shape sensing mechanism with a data-driven technique, that unlike conventional fiber shape sensors, does not include off-axis strain measurement and curvature calculation at discrete points along the fiber sensor's length to estimate its 3D shape. We used an easy-to-fabricate eFBG sensor with a simple and cost-effective readout unit. We designed a deep learning algorithm that can directly learn from our sensor's signal to predict its corresponding shape. We then evaluated the shape prediction accuracy of our designed model (the DL method) in various testing conditions and compared it with an exemplary experiment, the mode-field dislocation method (MFD). Furthermore, we showed that the spatial resolution of off-axis strain measurement in FBG-based (quasi-distributed) shape sensors is the main limitation, as the deformations between the sensing planes are not detected in complex shapes. The deep learning technique, on the other hand, uses the full spectrum of our eFBG sensor, including the Bragg resonance's side slopes, as the model's input to compensate for the low density of sensing planes. We believe that the deep learning model is using the impact of undesired bending-induced phenomena, including cladding mode coupling, bending-loss oscillations, and polarization-dependent losses, as additional sources of information to overcome the spatial resolution limitation for detecting complex deformations. Therefore, there is no need to adapt the fiber sensor design and its interrogation system for minimizing the impact of such bending-induced phenomena. The shape prediction error of our developed DL method for 3D curves in a curvature range of \SIrange{0.58}{33.5}{\m^{-1}} is reduced by a factor of $\sim 50$ compared to the MFD method. We also showed that the designed deep learning model generalizes nicely, as the performance is twice as good compared to a dictionary-based algorithm. The proposed shape sensing solution is 30 times less expensive than the commercially available distributed fiber shape sensor with a similar level of accuracy.


\section*{Methods}\label{}
\textbf{Working Principle of eFBG Sensor}
When the eFBG sensor is bent, the field distribution of the fundamental mode moves away from the core center~\cite{bao2018all, waltermann2018multiple,rong2017highly} (see Fig.~\ref{fig:mode_field}b). Dislocations in the mode-field's centroid cause intensity changes in the reflected signal from the eFBGs~\cite{waltermann2018multiple}. From the intensity ratio between the eFBGs at each sensing plane, curvature and bending direction can be calculated~\cite{waltermann2018multiple}. For simplification, this approach assumes that no other physical phenomena inside a bent optical fiber affect the intensity ratio between the eFBGs of the same sensing plane.

However, positioning FBGs away from the core axis breaks the cylindrical symmetry of the fiber, which increases coupling from the core mode to the cladding modes~\cite{thomas2011cladding,thomas2012cladding}.
The strength of such mode coupling varies when the fiber is bent, as it affects the overlap integral between the interacting modes~\cite{erdogan1997cladding,thomas2011cladding}. Bending an optical fiber causes strain-induced refractive index changes and dislocates the intensity distribution of the propagating light~\cite{shao2010directional,shao2010directional,bao2018all}, which directly influences the coupling efficiency. Therefore, the intensity of the cladding modes changes when the fiber is bent. In eFBGs, formation of cladding-mode resonances in fiber gratings provides highly sensitive full directional bending response with a simple light intensity measurement~\cite{feng2016off}. Although cladding modes are often stronger in stripped fibers or in fibers with lower refractive index coatings than the cladding layer~\cite{thomas2011cladding,thomas2012cladding}, they have also been observed in standard fibers coated with higher refractive index materials~\cite{renoirt2013high}.
Any recoupling between the excited cladding resonances and the fundamental mode, affects the relative intensity values between the eFBGs.

FBG interrogators for quasi-distributed sensors typically consist of a broadband light source, like a super luminescent diode (SLED), and a grating-based spectrometer. The emitted light from SLEDs is partially polarized, meaning that it undergoes wavelength-dependent polarization changes~\cite{galtarossa2005polarization} in a birefringence medium (\textit{e.g.,} bent fiber)~\cite{smith1980birefringence,ulrich1980bending,kersey1997fiber,block2006bending}. On the other hand, the efficiency of spectrometer grating is polarization-dependent, and therefore, the spectral profile will be impacted by polarization-dependent losses. This effect further modifies the measured intensity ratio between the Bragg peaks. The polarization effect in intensity-based fiber sensors is often kept at a minimum by using a polarization scrambler to change the polarization state randomly or by using polarization-insensitive spectroscopy instruments.

As is well known, light power loss increases when optical fibers bend~\cite{marcuse1976field}. Macro bending loss usually reflects itself in spectral modulations due to coherent coupling between the core mode and the radiated field reflected by the cladding-coating and the coating-air interfaces (also known as whispering gallery modes)~\cite{faustini1997bend, morgan1990wavelength}. The reflected field, at the coating-air boundary, causes short-period modulations as the re-injection path is longer~\cite{faustini1997bend, morgan1990wavelength}. Whereas, reflections at the closer cladding-coating interface cause long-period resonances~\cite{valiente1989new, harris1986bend, murakami1978bending, morgan1990wavelength}.
These bending attenuation losses are also temperature-dependent. Thermal variations affect the refractive index of the coating layer and consequently influence the coupling between the core and the cladding whispering gallery modes~\cite{morgan1990temperature}. Many models have been proposed to evaluate bending loss peak positions and shapes (\cite{valiente1989new, renner1992bending, harris1986bend}). The strong wavelength dependence of bending losses is an additional complicating factor in designing intensity-based sensors~\cite{morgan1990wavelength} as it modulates the spectral profile and affects the intensity ratio at the Bragg peaks of the eFBGs in a same sensing plane.\\
\\
\textbf{Setup}. 
\begin{figure}[!b]
    \centering
    \includegraphics[width=0.8\textwidth]{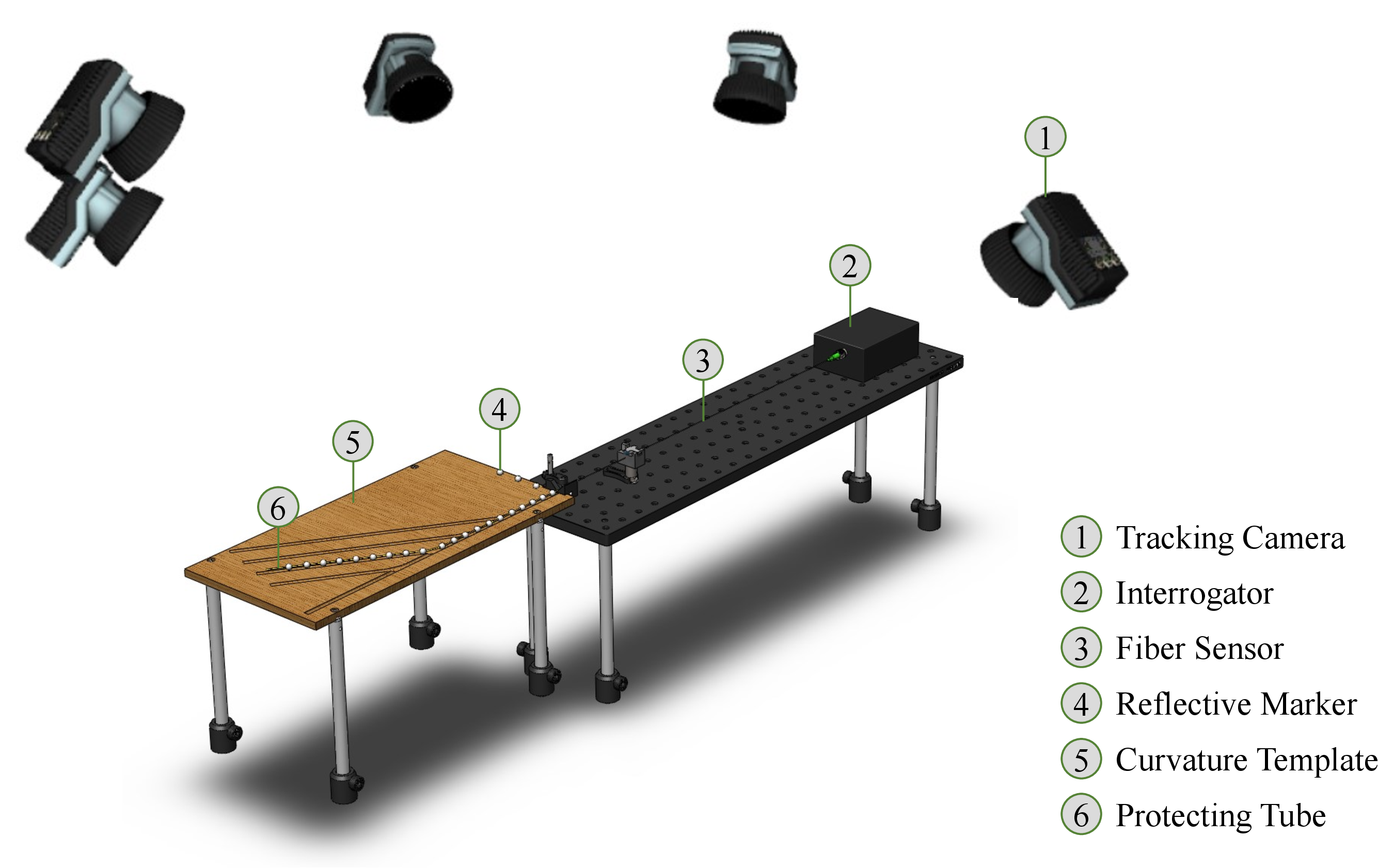}
    \caption{The data acquisition experimental setup. The motion capture system includes five tracking cameras (Oqus 7+, Qualisys AB, Sweden). For protection purposes, the fiber sensor is inserted in a Hytrel furcation tubing with an inner diameter of \SI{425}{\micro\m} and an outer diameter of \SI{900}{\micro\m}. Two v-clamps are used to hold the protection tubing and to fix the optical fiber before the insertion. The reflective markers are \SI{6.4}{mm} in diameter with an \SI{1}{mm} opening (X12Co., Ltd, Bulgaria). A thermocouple is placed close to the sensor's base to monitor the temperature during the data acquisition, ensuring no sudden thermal fluctuation affects the sensor's signal.}
    \label{fig:setup}
\end{figure}
Data acquisition setup used for developing the deep-learning-based model is shown in Fig.~\ref{fig:setup}. We used a low-cost FBG interrogator (MIOPAS GmbH, Goslar, Germany) consisting of an uncooled transmit optical sub-assembly (TOSA) SLED module and a NIR micro-spectrometer with \SI{0.5}{\nm} resolution to cover all 15 Bragg wavelengths from \SI{813}{nm} to \SI{869}{nm}. We recorded the sensor's spectra at random curvatures and orientations (in a curvature range of \SIrange{0.58}{33.5}{\m^{-1}}) while monitoring the reflective markers attached to the \SI{30}{\cm} long sensor using a motion capture system (Oqus 7+, Qualisys AB, Sweden). The data acquisition time period was 30\,minutes for $Test_1$ and 3\,minutes for $Test_2$ datasets. The acquisition rates in the FBG interrogator and the motion capture system were \SI{75}{\Hz} and \SI{200}{\Hz}, respectively. The sensor's spectra and the coordinate values corresponding to its shape were synchronized with a tolerance of less than \SI{3}{\ms}. 

We also used a laser-cut curvature template (Fig.~\ref{fig:setup}) to collect 320 samples for $Test_3$ dataset, when only certain sensor regions should be bent. The curvature template has four grooves allowing the sensor to be bent at the middle \SI{30}{\mm} area between the sensing planes 2 and 3, 3 and 4, 4 and 5, and \SI{10}{\mm} after the last sensing plane with a bending radius of \SI{50}{\mm}.\\
\\
\textbf{Training Setup}. 
The search space we defined for tuning the network's hyperparameters consists of the number of 1D convolutional layers (Conv1D), the number of fully connected layers (FC), the layer settings, the choice of batch normalization and downsampling, training settings, and loss function parameters. Search criteria are presented in Table~\ref{tab:SearchingCriterion}.

In the designed network, input samples with a batch size of 256 are first batch normalized and then fed into a Conv1D layer with 16 channels, followed by a max pooling layer with a kernel size of 3 and a stride of 2. The second Conv1D layer also has 16 channels, followed by a max pooling layer with a kernel size of 2. The third Conv1D layer has 32 channels, followed by a max pooling layer with a kernel size of 3 and a stride of 2. The fourth Conv1D layer also has 32 channels with a stride of 2, followed by a max pooling layer with a kernel size of 3. The last Conv1D layer has 256 channels, followed by batch normalization and a max pooling layer with a kernel size of 2 and a stride of 2. The extracted features are flattened to a 2048-long vector, fed into 5 FC layers, each with 2000 units. The first FC layer is followed by batch normalization and a dropout layer with a probability of 0.37, and two more FC layers. A batch normalization, an FC layer, a dropout layer with a probability of 0.16, and a fourth FC layer are the remaining layers before the final layer. The last layer is an FC layer that maps the output of the fourth FC layer into the target values, the relative coordinates. In all layers of this network architecture, the rectified linear unit (ReLU) serves as the activation function, and the kernel size for the Conv1D layers is 3. In this model, the Adam optimizer with a learning rate of 0.0001 minimizes the SmoothL1 loss function with a threshold of 4.04. \\
\begin{table}[!t]
    \fontsize{8}{10}\selectfont
    \caption{Search criteria for hyperparameter optimization.}
    \label{tab:SearchingCriterion}
    \begin{tabular}{lll}
    \hline
    \textbf{hyperparameter} & \textbf{search space} & \textbf{selected values} \\ \hline
    number of Conv1D layer    & min: 1, max: 20, step: 1 & 5 \\ 
    number of FC layer  & min: 1, max: 20, step: 1 & 5 \\
    BN after each layer & true, false & -\\
    dropout after FC layer & true, false & -\\
    dropout rate & min: 0.1, max: 0.8 & -\\
    stride  & min: 1, max: 2, step: 1  & - \\ 
    kernel size (max pooling layer)    & min: 2, max: 3, step: 1  & - \\
    distribution of initial weights & \begin{tabular}[l]{@{}l@{}}standard, Xavier\_uniform, \\ Xavier\_normal, Kaiming\_uniform,\\ Kaiming\_normal\end{tabular} &  Xavier\_normal\\
    learning rate   & 0.01, 0.001, 0.0001, 0.00001  &  0.0001\\ 
    sorting Conv1D layers & true, false   & true \\
    L2 regularization   & 0.1, 0.01, 0.001, 0.0001, 0.00001, 0  & 0 \\
    threshold in SmoothL1 & any values between 0.0 and 5.0    &  4.04 \\ \hline
    \end{tabular}
\end{table}
\\
\textbf{Decoding The Model's Decisions}. 
Inspired by Gradient-weighted Class Activation Mapping (Grad-CAM), we decode the decisions made by our CNN (convolutional neural network)-based model. Decoding our model's decisions helps us understand which part of the input spectra contributes to coordinate predictions. Grad-CAM is a commonly used technique in image classification problems that generates visual explanations from any CNN-based model without any re-training or architectural changes required. Gradient is a measure that shows the effect on the output caused by the input. In other words, we are looking for the part of the input with the highest effect on the model's output. However, due to the small output dimension in each channel of the last Conv1D layer, its gradient heat map highlights the inputs' important parts with a low resolution. Therefore, instead of the gradient of the Conv1D layers, we calculate the forward finite difference of the model's loss with respect to the input spectral elements. The spacing constant is chosen $0.1$, higher than the spectral intensity noise level. In this method, we modify the intensity value of one spectral element and monitor the changes of the model's loss value. We repeat this process for all 190 spectral elements. The resultant color maps (shown in Fig.~\ref{fig:750}~(b) and Fig.~\ref{fig:1320}~(b)) indicate the impact of the changes in each spectral element on the model's SmoothL1 loss value. In order to investigate the contribution of each spectral element to the coordinate prediction of each individual marker, we calculated the Euclidean distance between the predicted coordinates of each marker before and after spectral modification. This way, we were able to highlight all the spectral elements contributing to the relative coordinate prediction of each marker. 

\bmhead{Supplementary Information}
We provided three videos in the supplementary material, visualizing the sensor's predicted shapes using the DL and the MFD methods on all three datasets.
\bmhead{Acknowledgments}
We gratefully acknowledge the funding of this work by the Werner Siemens
Foundation through the MIRACLE project.

\bibliography{mainbody}

\end{document}